# A Novel Progressive Multi-label Classifier for Class-incremental Data

Mihika Dave*, Sahil Tapiawala
Dept. of Electrical & Electronics Engineering
Birla Institute of Technology & Science (BITS)-Pilani
Goa, India

Meng Joo Er, Rajasekar Venkatesan
School of Electrical & Electronics Engineering
Nanyang Technological University
Singapore

*Abstract*— **In this paper, a progressive learning algorithm for multi-label classification to learn new labels while retaining the knowledge of previous labels is designed. New output neurons corresponding to new labels are added and the neural network connections and parameters are automatically restructured as if the label has been introduced from the beginning. This work is the first of the kind in multi-label classifier for class-incremental learning. It is useful for real-world applications in applied fields such as robotics where streaming data are available and the number of labels is often unknown. Based on the Extreme Learning Machine framework, a novel universal classifier with plug and play capabilities for progressive multi-label classification is developed. Experimental results on various benchmark synthetic and real datasets validate the efficiency and effectiveness of our proposed algorithm.**

## I. INTRODUCTION

In the modern context of machine intelligence, the growing importance of classification has motivated the development of several algorithms which are scalable [1][2]. A number of tasks right from the identification of objects in images to the study of emotion-associated brainwaves belong to this category. Classification, which is the identification of the target categories a data sample belongs to, can be divided into single-label and multi-label classification. Single-label classification involves binary and multi-class classification where a data sample is associated with one label only. On the contrary, multi-label classification involves data samples which are simultaneously associated with multiple labels. The learning algorithms are generally of two types: Batch learning and Sequential learning. Batch learning requires pre-collection of training data, and the parameters of the network are calculated by processing all the training data concurrently. While, in online/sequential learning algorithms, the network parameters are updated as and when a new training data arrives [3], [4].

Multi-label classification has become significant due to its rapidly increasing application areas such as text categorization [6]-[8], bioinformatics [9], [10], medical diagnosis [11], scene classification [12], genomics, map labeling [13], marketing, multimedia, music categorization, etc. The rising significance of multi-label classification has spurred a recent growth in its theoretical analysis [14], [15] and development of algorithms for practical applications [16], [17].

The existing multi-label classifiers once trained to classify a specific number of labels, cannot learn new labels without retraining all the labels anew again. Such classifiers work well with the pre-known dataset, but they may not be appropriate for applications such as cognitive robotics or real-time applications of big data where the nature of training data is unknown. For such real-time data, the learning technique must be self-adapting to suit the dynamic needs. Class-incremental Extreme Learning Machines (ELM) has been proposed for multi-class classification [18] but there is no significant work done for multi-label classification. To overcome this shortcoming, a novel learning paradigm based on Extreme Learning Machine is proposed, called the "*progressive-ELM multi-label classifier*" (Pro-EMLC). This is the first method for incremental learning in multi-label classification. It has been successfully tested on benchmark datasets like Scene, Corel5k, and Medical. The promising results we obtained validate the efficacy of our approach. Next section gives a brief description of ELM and Online Sequential-ELM. Section 3 describes the proposed method. Section 4 presents the results obtained and section 5 states the conclusion.

## II. BRIEF OVERVIEW OF ELM AND OS-ELM

ELM considers a 'generalized' Single Hidden Layer Feedforward Neural Network (SLFN) architecture consisting of n input neurons and m output neurons, with N hidden layer neurons in the second layer. Most neural networks are considered to be universal classifiers or function approximators [19], [20] when all the parameters of the neural network are tuned. It has been previously shown that ELM works for the SLFN architecture without tuning the hidden layer parameters (feature mapping parameters) [21]. We discuss the ELM paradigm in brief in the following paragraphs.

Consider there are *N'* hidden layer neurons, *n* is the number of features and *m* is the number of labels. The training data of size N samples is of the form $\{(x_i, y_i) | x_i \in R^n, y_i \in R^m, i = 1,...N\}$, where $x_i$ represents input feature vector and $y_i$ represents the output label vector. Label space L = $\{Y_1, Y_2,...,Y_M\}$. The predicted output of SLFN '$o_j$' for *j = 1,...., N* is:

$$\sum_{i=1}^{N'} \beta_i g(w_i \cdot x_j + b_i) = o_j \quad (1)$$

Where, *g(x)* is the activation function, $w_i = [w_{i1}, w_{i2}, ..., w_{in}]^T$ is the input weight, $\beta_i = [\beta_{i1}, \beta_{i2},...\beta_{im}]^T$ is the output weight, and $b_i$ is the hidden layer bias.

In ELM, the input weights and hidden layer bias are assigned randomly. To minimize the difference between the actual and predicted output, $\beta_i$ should be found such that output class is equal to the target class.

$$\sum_{j=1}^{N'} \|o_j - y_j\| = 0 \quad (2)$$

Thus, ELM output network can be written as

$$\sum_{i=1}^{N'} \beta_i g(w_i \cdot x_j + b_i) = y_j, \ j=1,\ldots, N \quad (3)$$

In matrix form,

$$H\beta = Y \quad (4)$$

Where,

$$H = \begin{bmatrix} g(w_1 \cdot x_1 + b_1) & \cdots & g(w_{N'} \cdot x_1 + b_{N'}) \\ \vdots & \ddots & \vdots \\ g(w_1 \cdot x_N + b_1) & \cdots & g(w_{N'} \cdot x_N + b_{N'}) \end{bmatrix}_{N \times N'}$$

$$\beta = \begin{bmatrix} \beta_1^T \\ \vdots \\ \beta_{N'}^T \end{bmatrix}_{N' \times m} \text{ and } Y = \begin{bmatrix} y_1^T \\ \vdots \\ y_N^T \end{bmatrix}_{N \times m}$$

Using the Moore-Penrose generalized inverse ($H^+$) of hidden layer matrix $H$, we can get the output feature mapping matrix $\beta$ of the ELM network.

$$\beta = H^+ Y \quad (5)$$

Where, $H^+ = (H^T H)^{-1} H^T$

In real time applications, the complete training data is seldom available and as the data arrives sequentially, the ELM has to be retrained with a combination of the new data and the previously available data. There is no provision to retain the learning from the previous data and train the network for only the new data. On-line Sequential Extreme Learning Machine (OS-ELM) retains the knowledge of the previous training data and can learn data over previously available data chunk-by-chunk with varying chunk sizes [22]. The above-mentioned algorithm of ELM is extended to OS-ELM as described below.

Let the initial block of training data have $N_0$ samples. For the initial block, compute

$$M_0 = (H_0^T H_0)^{-1} \quad (6)$$

$$\beta_0 = M_0 H_0^T Y_0 \quad (7)$$

Where $H_0 = [h_1 \ldots h_{N0}]^T$ is the feature mapping (hidden layer) matrix computed as above and $Y_0 = [y_1,\ldots,y_{N0}]^T$ output vector for $N_0$ samples

For the incoming $(k+1)^{th}$ sequential data, Recursive Least Squares (RLS) approximation is used to retain the learning. Updating can be done as,

$$M_{k+1} = M_k - \frac{M_k h_{k+1} h_{k+1}^T M_k}{1 + h_{k+1}^T M_k h_{k+1}} \quad (8)$$

$$\beta_{k+1} = \beta_k + M_{k+1} h_{k+1}(Y_{k+1}^T - h_{k+1}^T \beta_k) \quad (9)$$

Where k = 0, 1, 2,.., N-$N_0$-1.

The calculated value of $\beta$ is used for predicting the output matrix. The detailed theory and mathematics involved in ELM and OS-ELM are given in [22] [23].

## III. PROPOSED METHOD

The steps of the proposed 'progressive-ELM multi-label classifier' are:

*1) Processing of input:* The raw input data is processed so that the output label, corresponding to each input sample, is an m-tuple with -1 or 1 representing the belongingness to each of the labels in the label space L.

*2) Initialization:* The input weights and the hidden layer bias are assigned at random in accordance with the ELM learning paradigm. The number of hidden layer neurons $N'$ is fixed such that over-fitting does not occur. The optimal value $N'$ is found by carrying out different epochs with varying number of hidden neurons, plotting it *w.r.t.* the training and cross-validation accuracy and choosing the optimal number of hidden neurons.

*3) ELM training – Initial learning:* The hidden layer output matrix $H_0$ is calculated for an initial block of $N_0$ data samples.

$$H_0 = [h_1,\ldots.h_{N0}]^T \quad (10)$$

Where $h_i = [g(w_1.x_i+b_1),\ldots.g(w_N:x_i+b_{N'})]^T$, i = 1,2…$N_0$.

Using $H_0$, the initial values of $M_0$ and $\beta_0$ are estimated as

$$M_0 = (H_0^T H_0)^{-1} \quad (11)$$

$$\beta_0 = M_0 H_0^T Y_0 \quad (12)$$

*4) ELM training – Sequential Learning:* The subsequent data arriving at the network can be trained one-by-one or chunk-by-chunk. Let the chunk size be 'b'. For b=1, data is trained one-by-one. The incoming data may have the presence of a new label(s), indicated by the increase in tuple size. Presence/Absence of a new label is handled as below:

*a) Absence:* In the absence of a new label, no special computations are required and step 5 is executed directly.

*b) Presence:* In the presence of a new label, the network is to be recalibrated to accommodate a new label, while retaining the old knowledge. Let 'c' new labels be introduced and m labels of data are currently learnt. Introducing 'c' new labels at any instant k+1 modifies the dimensions of the output weight matrix $\beta_{N'Xm}$ to $\beta_{N'Xm+c}$. Transformed output weight matrix $\widetilde{\beta_k}$ is given as $\widetilde{\beta_k} = (\beta_k)_{N'Xm} \ I_{mXm+c}$ where, $I_{mXm+c}$ is m X m+c dimensional rectangular identity matrix.

$$\widetilde{\beta_k}_{N'Xm+c} = (\beta_k)_{N'Xm} \begin{bmatrix} 1 & 0 & \cdots & 0 \\ 0 & 1 & \cdots & 0 \\ 0 & 0 & \cdots & 0 \\ 0 & 0 & \cdots & 0 \end{bmatrix}_{mXm+c} \quad (13)$$

$$\widetilde{\beta_k}_{N'Xm+c} = [(\beta_k)_{N'Xm} \ O_{N'Xc}]_{N'Xm+c} \quad (14)$$

where $O_{N'Xc}$ is a zero matrix.

The output values corresponding to new labels is -1 for previously learnt samples. Therefore, the k-learning step update for the 'c' new labels $((\Delta\beta_k)_{N'Xc}(\Delta\beta_k)_{N'Xc})$ can be expressed as,

$$(\Delta\beta_k)_{N'Xc} = (M_k)_{N'XN'} (h_k^T)_{N'Xb} \begin{bmatrix} -1 & \cdots & -1 \\ \vdots & \ddots & \vdots \\ -1 & \cdots & -1 \end{bmatrix}_{bXc} \quad (15)$$

$$(\Delta\beta_k)_{N'Xc} = -(M_k)_{N'XN'} (h_k^T)_{N'Xb} J_{bXc} \quad (16)$$

Where, $J_{bXc}$ is an all-ones matrix of size b x c.

$$(\Delta\beta_k)_{N'Xm+c} = [O_{N'Xm} \quad -(M_k)_{N'XN'} (h_k^T)_{N'Xb} J_{bXc}] \quad (17)$$

The recalibrated output weight matrix $(\beta_k)_{N'X(m+c)}$ is calculated as,

$$(\beta_k)_{N'X(m+c)} = \widetilde{\beta_k}_{N'Xm+c} + (\Delta\tilde\beta_k)_{N'Xm+c} \quad (18)$$

Upon simplification, $(\beta_k)_{N'X(N'+c)}$ can be expressed as,

$$(\beta_k)_{N'X(m+c)} = [(\beta_k)_{N'Xm} \quad (\Delta\beta_k)_{N'Xc}] \quad (19)$$

The hidden layer output vector $h_{k+1}$ is calculated. The output weight is updated according to the Recursive Least Square algorithm [22],

$$M_{k+1} = M_k - M_k h_{k+1}^T (I + h_{k+1} M_k h_{k+1}^T)^{-1} h_{k+1} M_k \quad (20)$$

$$\beta_{k+1} = \beta_k + M_{k+1} h_{k+1} (Y_{k+1}^T - h_{k+1}^T \beta_k) \quad (21)$$

*5) ELM testing and thresholding:* Raw output matrix $Y$ of the testing data samples is calculated *using $Y = H\beta$*. Since the number of labels a samples belongs to is unknown, thresholding of raw output is suggested. Thresholding refers to the application of a threshold value, based on the separation between two categories of labels (Labels that the data sample belongs to and labels the data sample does not belong to), to identify the number of labels and the target class labels corresponding to the input data sample. As a trivial case, zero threshold is chosen. Passing the raw output through bipolar step function gives the samples' association to the label(s).

Thus, the proposed algorithm allows for class-incremental training of multi-label data.

## IV. EXPERIMENTAL RESULTS

The proposed method is tested on benchmark datasets, namely Scene, Corel5k and Medical, having a varied number of labels (6 to 374) and belonging to diverse domains like multimedia and text. The degree of 'multi-labelness' of datasets is measured by label density (LD) and label cardinality (LC) as defined by Tsoumakas et al [24]. Label density of the tested datasets ranges from 0.009 to 0.178 and label cardinality ranges from 1.07 to 3.52. The characteristics of these datasets are given in Table I. To perform testing for progressive learning, the data samples are redistributed such that the number of labels present in the initial block of data is less than the total number of labels present in the dataset. New labels can be introduced in the streaming data one-by-one or in groups.

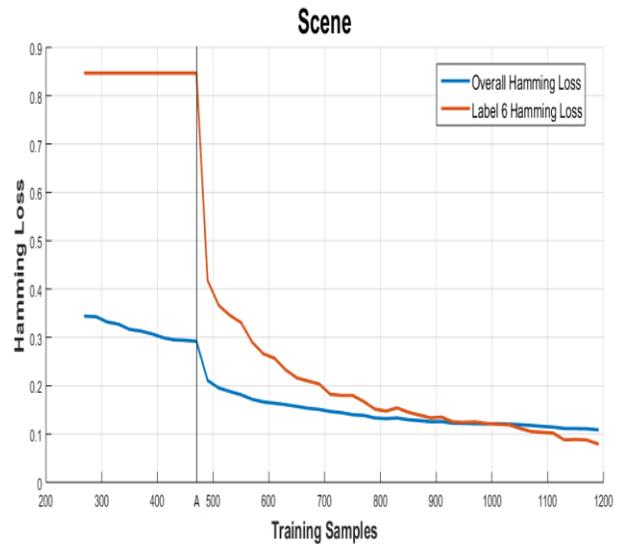

Fig. 1. Hamming Loss for Scene Dataset

TABLE I. CHARACTERISTICS OF DATASET

| Characteristic | Dataset | | |
| --- | --- | --- | --- |
| | *Scene* | *Corel5k* | *Medical* |
| Domain | Multimedia | Multimedia | Text |
| No. of features | 294 | 499 | 1449 |
| No. of samples | 2407 | 5000 | 978 |
| No. of labels | 6 | 374 | 45 |
| Label Density | 0.178 | 0.009 | 0.027 |
| Label Cardinality | 1.07 | 3.52 | 1.25 |

The learning curve or the hamming loss is plotted for Scene dataset (having 6 labels) in Fig. 1. The hamming loss decreases with increasing number of samples. Till point A, when only five

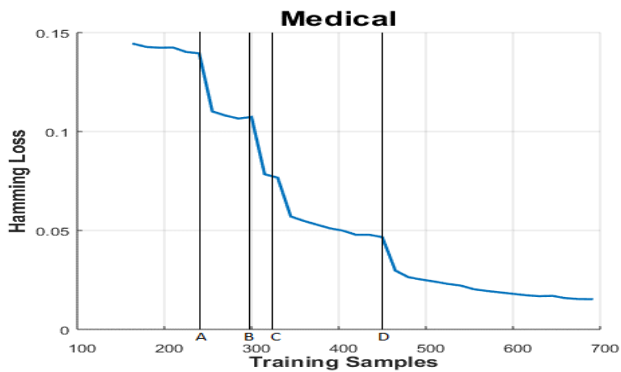 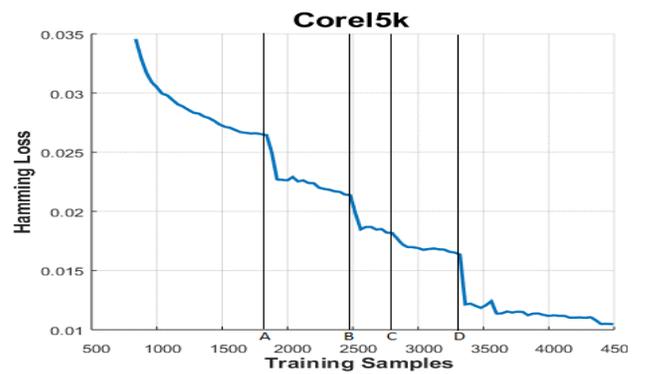

Fig. 2. Hamming loss for medical (39+2+2+1+1) and corel5k (368+2+1+1+2) dataset

TABLE II. PERFORMANCE METRIC

| Data set | LIP | H | Acc | Pre | Rec | F1 | T1 | T2 |
|---|---|---|---|---|---|---|---|---|
| Scene | 5+1 | 0.104 | 0.609 | 0.627 | 0.659 | 0.643 | 2.266 | 0.042 |
| Scene | 4+1+1 | 0.139 | 0.569 | 0.584 | 0.699 | 0.636 | 2.291 | 0.080 |
| Medical | 44+1 | 0.012 | 0.693 | 0.740 | 0.729 | 0.734 | 0.604 | 0.025 |
| Medical | 39+2+2+1+1 | 0.016 | 0.653 | 0.695 | 0.731 | 0.712 | 0.611 | 0.034 |
| Medical | 38+3+4 | 0.023 | 0.585 | 0.622 | 0.739 | 0.675 | 0.614 | 0.027 |
| Corel5k | 373+1 | 0.010 | 0.057 | 0.164 | 0.059 | 0.087 | 5.348 | 0.044 |
| Corel5k | 368+2+1+1+2 | 0.010 | 0.055 | 0.151 | 0.062 | 0.088 | 5.396 | 0.056 |

(LIP- Label Introduction Pattern, H- Hamming Loss, Acc- Accuracy, Pre- Precision, Rec- Recall, F1- F1 Score, T1- Train Time, T2- Test Time)

labels are learnt, the overall hamming loss is higher since the prediction for the 6$^{th}$ label is incorrect. After introducing the remaining label (label 6), at point A, in the sequential phase, there is a sharp decrease in hamming loss suggesting that the prediction for label 6 has improved. The hamming loss for the label 6 alone is also shown in Fig. 1. The learning curves for Medical and Corel5k datasets are shown in Fig. 2. For the Medical dataset, 39 out of 45 labels were introduced initially. To verify the behavior towards incremental labels, two labels are introduced at point A, two labels at point B, one label at point C, and finally the last label at point D for the Medical dataset. The falling hamming loss is evidence of the improvement in the learning resulting in better prediction. This pattern of introducing new labels is represented as 39+2+2+1+1 in Table II. Similar protocol was followed for Corel5k dataset with the introduction pattern being 368+2+1+1+2.

TABLE III. STATE-OF-THE-ART MULTI-LABEL CLASSIFIERS

| Method Name | Category |
|---|---|
| Classifier Chain (CC) | SVM |
| QWeighted approach for Multi-label Learning (QWML) | SVM |
| Hierarchy Of Multi-label ClassifiERs (HOMER) | SVM |
| Multi-Label C4.5 (ML-C4.5) | Decision Trees |
| Predictive Clustering Trees (PCT) | Decision Trees |
| Multi-Label k-Nearest Neighbors (ML-kNN) | Nearest Neighbors |
| Ensemble of Classifier Chains (ECC) | SVM |
| Random Forest Predictive Clustering Trees (RF-PCT) | Decision Trees |
| Random Forest of ML-C4.5 (RFML-C4.5) | Decision Trees |

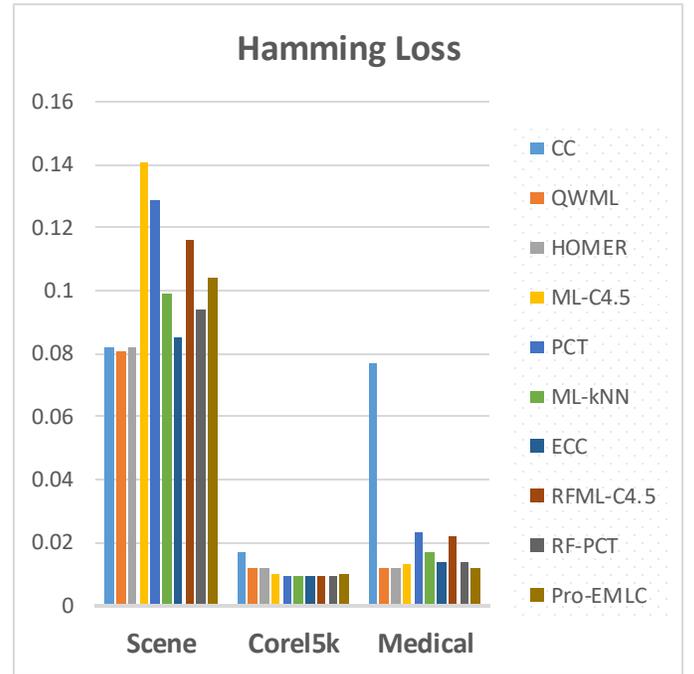

Fig. 3. Comparison of hamming loss with other multi-label classifiers

A 10-fold cross-validation is carried out for various combinations of label introduction in Scene (6 labels), Medical (45 labels) and Corel5k (374 labels) data sets. Various measures to evaluate the performance of multi-label classifiers, as explained by Tsoukamas et al [24] are calculated for the proposed algorithm. The results are presented in Table II. Label introduction pattern is the number of labels introduced in the initial phase followed by the number of those in the sequential phase.

The proposed algorithm was compared to the state-of-the-art online learning multi-label algorithms mentioned in Table III.

TABLE IV. COMPARISION OF TRAINING TIME FOR VARIOUS MULTI-LABEL CLASSIFIERS

| Classifier | Dataset | | |
|---|---|---|---|
| | *Scene* | *Medical* | *Corel5k* |
| CC | 99 | 1125 | 28 |
| QWML | 195 | 2388 | 40 |
| HOMER | 68 | 771 | 16 |
| ML-C4.5 | 8 | 369 | 3 |
| PCT | 2 | 30 | 0.6 |
| ML-kNN | 14 | 389 | 1 |
| ECC | 319 | 10073 | 103 |
| RFML-C4.5 | 10 | 385 | 7 |
| RF-PCT | 23 | 902 | 27 |
| Pro-EMLC | 2.266 | 5.396 | 0.611 |

Despite being an online-class incremental algorithm, the hamming loss of the proposed Progressive - ELM Multi-label Classifier (Pro-EMLC), as shown in Fig. 3, was comparable to the state-of-the-art multi-label classification techniques for batch learning described by Madjarov et al. [25], for all the tested datasets. A similar comparision of training time was also made as given in Table IV. Since the algorithm is based on ELM, the training and testing speed is very high, making it highly suitable for real-time applications of big data analysis.This suggests that the performance of the algorithm is superior for a class incremental algorithm and hence can be used widely.

V. CONCLUSIONS

The proposed progressive-ELM multi-label classifier is the first of its kind. It can be used for online/streaming big data applications with known number of labels as well as for real-time applications such as cognitive robotics where the number of labels is unknown. It has shown high speed and high performance metric for all the 3 tested benchmark datasets. Based on these promising results, the proposed ELM-based classifier can be considered fit for achieving high performance with speed for multi-label classification, especially in incremental learning areas.

ACKNOWLEDGEMENT

The authors would like to acknowledge the funding support from the Ministry of Education, Singapore (Tier 1 AcRF, RG30/14).